\documentclass[sigconf, screen]{acmart}
\AtBeginDocument{%
  }

\setcopyright{acmlicensed}
\copyrightyear{2018}
\acmYear{2018}
\acmDOI{XXXXXXX.XXXXXXX}
\acmConference[Conference acronym 'XX]{Make sure to enter the correct
  conference title from your rights confirmation email}{June 03--05,
  2018}{Woodstock, NY}
\acmISBN{978-1-4503-XXXX-X/2018/06}

\usepackage{enumitem}
\setlist[itemize]{leftmargin=*}
\begin{document}

\title{Child-Oriented AIGC Video Risk Reviewing: A Benchmark and Knowledge-Supported Iterative Reasoning Framework}

\author{Lewen Mi}
\affiliation{
  \institution{Shandong University}
  \city{Weihai}
  \state{Shandong}
  \country{China}
}

\author{Manyi Li}
\affiliation{
  \institution{Shandong University}
  \city{Jinan}
  \state{Shandong}
  \country{China}
}

\author{Yuling Sun}
\affiliation{
  \institution{Fudan University}
  \city{Shanghai}
  \country{China}
}

\author{Yufan Zhang}
\affiliation{
  \institution{Shandong University}
  \city{Weihai}
  \state{Shandong}
  \country{China}
}

\author{Yuxin Shi}
\affiliation{
  \institution{Beijing Jiaotong University}
  \city{Weihai}
  \state{Shandong}
  \country{China}
}

\author{Yulong Bian}
\affiliation{
  \institution{Shandong University}
  \city{Weihai}
  \state{Shandong}
  \country{China}
}

\author{Xiangxian Li}
\affiliation{
  \institution{Shandong University}
  \city{Weihai}
  \state{Shandong}
  \country{China}
}

\author{Juan Liu}
\affiliation{
  \institution{Shandong University}
  \city{Weihai}
  \state{Shandong}
  \country{China}
}

\renewcommand{\shortauthors}{Mi et al.}



\begin{abstract}

The rapid growth of Artificial Intelligence-generated content (AIGC) is reshaping video production and circulation, exposing children to an increasing volume of AIGC videos. Unlike traditionally produced videos, AIGC videos often exhibit greater uncertainty in visual details, narrative coherence, and content expression, which may introduce developmentally inappropriate risks for children.However, existing video safety research is largely designed for general violation detection from an adult perspective and remains insufficient for identifying the fine-grained, implicit, and context-dependent risks that children may encounter when viewing AIGC videos.To address this gap, we study child-oriented AIGC video reviewing, making three contributions. First, we construct CAVSR, a benchmark of 605 real-world videos collected from multiple platforms, and develop a hierarchical risk taxonomy comprising 6 top-level categories and 26 fine-grained labels to support systematic evaluation of children’s viewing risks. Second, we propose QVRS-E, a knowledge- and experience-augmented video reviewing framework that combines multi-agent collaboration with expert and experiential knowledge to support targeted evidence acquisition and fact-grounded reviewing decisions. Third, extensive experiments demonstrate that our method significantly enhances the reviewing of child-related risks integrated with vision-language models, and yields more robust review reports.

\end{abstract}



\begin{teaserfigure}
  \includegraphics[width=\textwidth]{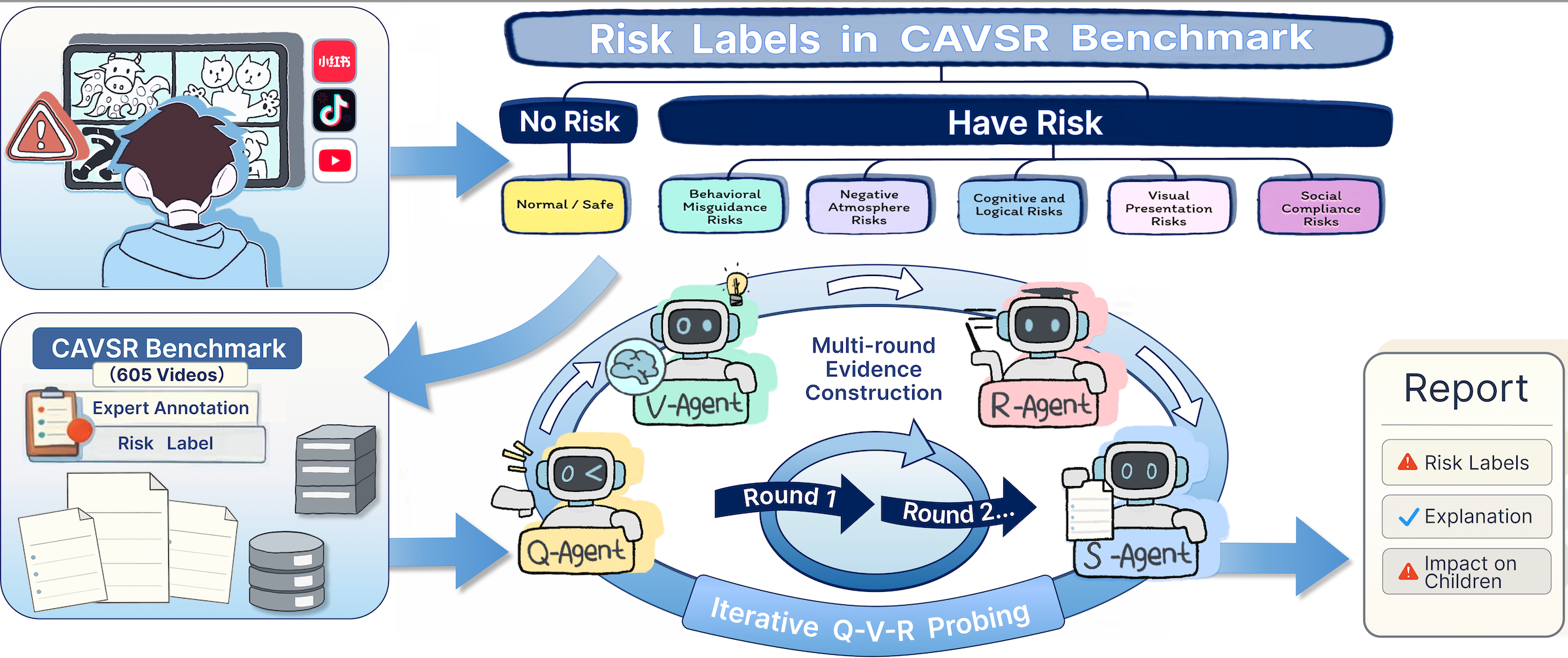} 
  \caption{Overview of our study on child-oriented AIGC video risk review. The figure illustrates the CAVSR benchmark, its risk taxonomy, and the proposed QVRS-E framework for iterative evidence construction and structured risk assessment.}
  \Description{A bar chart showing the count of different video categories in the database.}
  \label{fig:int}
\end{teaserfigure}

\received{20 February 2007}
\received[revised]{12 March 2009}
\received[accepted]{5 June 2009}

\maketitle
\section{Introduction}

The rapid development of Artificial Intelligence-Generated Content (AIGC) technologies is reshaping how video content is produced and circulated. AIGC-generated videos are now entering mainstream content production and consumption at an unprecedented pace, spreading rapidly across short-video platforms and open media environments \cite{chen2025ai,huang2023recent,orak2024using}. Children\footnote{{In this paper, ``children'' mainly refers to preschool-aged children (approximately 3-6 years old). }}, as one major audience on these platforms \cite{bozzola2022use,hadjipanayis2019social}, are increasingly exposed to such content. Yet, compared with traditionally produced videos, AIGC videos often involve greater uncertainty in visual details, narrative logic, and content expression \cite{cao2026failures,chu2026evaluation}, and are more likely to contain problems such as distorted content, incoherent logic, excessive emotional rendering, and ambiguous value messages \cite{wrro223407,qi2024sniffer,ramesh2022beach,wei2025contextaware}.
For children, whose cognitive and judgment abilities are still developing, any inappropriate character design, depictions of dangerous actions, distorted knowledge, emotionally exaggerated atmospheres, or biased value messages embedded in such video content may negatively influence their emotional experience, behavioral imitation, cognitive development, and value formation
\cite{yousaf2022deep}. 
With rising child exposure to AIGC videos, targeted safety reviewing for child-specific risks has become an urgent priority.

Technically, child-oriented video reviewing is not equivalent to traditional unsafe video content detection, whose primary goal is to determine whether a video contains clearly prohibited elements, such as violence, pornography, and bloodshed \cite{kollias2025dvd,azim2026explainable}. 
Instead, it must go beyond identifying conventional violations and further assess whether a video's style of expression, narrative construction, and affective atmosphere may cause harmful or developmentally inappropriate effects on children \cite{bonus2025you}. Importantly, many of these risks are not unique to AIGC videos, but are further amplified by limited controllability in generation, weak cross-frame consistency \cite{ma2025controllable,qu2024exploring,zhou2026semantic}, and blurred expressive boundaries of AIGC, making the identification of child-related risks substantially more complex.

Existing research still struggles to effectively support this task. On the one hand, most current datasets (e.g. \cite{singh2019kidsguard,liu2025video,wang2025safevid}) are designed for general violation detection or broad content safety assessment, making them unsuitable for fine-grained risk evaluation from the perspective of children's potential exposure and impact. On the other hand, existing methods typically rely on single-pass captions, holistic classification, or short-context judgments (e.g. \cite{omarov2022state,khan2025violence,huszar2023toward}), which are better suited to identifying visually salient and explicit risks, but often insufficient for detecting more implicit risks that are distributed across multiple frames and require step-by-step verification through contextual reasoning. 

To address this gap, this paper investigates the reviewing of AIGC videos with respect to children's viewing risks, as illustrated in Figure~\ref{fig:int}.
We first construct a video benchmark,
named CAVSR, consisting of 605 videos collected from real-world platforms, including TikTok, RedNote, and YouTube. 
Rather than determining whether a video is intended for children, the annotation focuses on identifying the potential risks children may face after watching it.
To support systematic evaluation, we developed a hierarchical risk taxonomy through expert focus group discussions. The taxonomy includes 6 primary risk categories and 26 fine-grained labels \cite{chand2025methods}. During its development, experts iteratively defined the risk categories, refined the label boundaries, and gradually reached consensus on the final taxonomy. Based on this finalized taxonomy, experts then annotated the benchmark data accordingly.

Building on this benchmark, we further propose QVRS-E, a knowledge-supported AIGC video reviewing framework that incorporates a multi-agent expert collaboration and consultation mechanism for better understanding implicit risks embedded in video content. When reviewing videos, the Question Agent (Q-Agent), Vision Agent (V-Agent), and Review Agent (R-Agent) collaborate over multiple rounds to iteratively collect and verify visual evidence, while the Summary Agent (S-Agent) integrates expert knowledge and experience knowledge to produce structured risk assessment results.
Specifically, the expert knowledge base stores rule-oriented knowledge for child content reviewing, including risk definitions, decision boundaries, typical manifestations, and corresponding reviewing considerations, thereby providing explicit guidance for question generation and risk assessment; the experience knowledge base stores scene descriptions, effective inquiry paths, and associated risk types from historical reviewing cases, offering transferable questioning strategies and attention cues for the current video.
Unlike approaches that rely on a single-step prediction, QVRS-E accomplishes the review decision through a series of questions, verifications, and evidence supplementation. This enables multiple rounds of evidence construction to continuously focus on key risk clues with the joint support of expert knowledge and experience knowledge. As a result, it helps reduce the possible omissions and hallucinations that may occur in single-step prediction, while enhancing the reliability of fine-grained risk judgment.

Experimental results show that the proposed framework consistently outperforms multiple baselines across both the 26 fine-grained and merged 6-class settings. The results suggest that iterative Q-V-R probing can effectively uncover risk-related evidence, while the incorporation of expert and experience knowledge further improves fine-grained risk attribution and label ranking. Moreover, when integrating with general-purpose open-source VLMs, our framework brings substantial performance improvements over their original backbones.

In summary, this paper makes the following three contributions:
\begin{itemize}
    \item We construct a child-oriented AIGC video benchmark, named CAVSR, based on 605 real-world videos collected from online platforms, and establish a hierarchical risk taxonomy with 6 top-level risk categories and 26 fine-grained labels, providing a systematic benchmark for evaluating children’s viewing risks.
    \item We propose a knowledge-supported AIGC video reviewing framework that performs more reliable risk recognition and structured risk assessment through iterative Q-V-R probing evidence construction, expert knowledge support, and experience update.
    \item We provide extensive empirical analysis showing that iterative Q-V-R probing mainly improves evidence coverage, while expert knowledge support further enhances the reliability of fine-grained child-risk recognition.
\end{itemize}

\section{Related Work}
\subsection{Child-Oriented Video Review}

Video review has been widely studied in various vertical fields. Significant work has focused on identifying harmful videos in adult settings, such as the detection of false video information \cite{jo2024harmful,qi2024sniffer}, multilingual hate video \cite{wang2024multihateclip}, and discrimination against disabled and marginalized groups \cite{lu2025mv,kojah2025silencing}. Moreover, most existing video dataset evaluations and benchmark tests are constructed from an adult perspective, emphasizing risks such as politically sensitive, misinformation, and explicit violence \cite{jo2024harmful,pramanick2021momenta,edstedt2022vidharm,singhal2023sok}, etc. In contrast, research on video review for children remains limited. 
Existing review systems have been shown to be vulnerable when dealing with child-related videos \cite{eltaher2025protecting,aggarwal2023protecting}. For example, malicious users often disguise adult content as children's cartoon videos for dissemination. Such content can easily bypass traditional keyword or metadata filters \cite{gkolemi2022youtubers,aggarwal2023protecting}, while the unsafe transcription illusion produced by automatic speech recognition (ASR) systems when processing children's videos further exacerbates the risk \cite{ramesh2022beach}. Although some studies have begun to explore children's video review through audio-visual fusion \cite{ahmed2024enhanced,ahmed2024multimodal}, deep recognition of cartoon character features \cite{yousaf2022deep,chuttur2022multi}, and age-adaptive detection models \cite{alam2024utilizing}, these methods remain largely oriented toward explicit content detection and are still insufficient for identifying the diverse, implicit, and developmentally situated risks that children may encounter when viewing videos \cite{steen2023you,ahmed2023potential,ahmed2023malicious,balat2024tikguard}.



\subsection{Multi-Agent Technologies for Video Review}
Video review has evolved from metadata-based filtering to multimodal recognition frameworks \cite{tang2021videomoderator,ami2021ai,binh2022samba}. With the rise of large language models (LLMs) and visual large language models (VLMs), researchers have increasingly explored their use for fine-grained and context-dependent harmful content detection \cite{jo2024harmful,kumar2024watch,levi2025ai,wei2025contextaware}. However, mainstream VideoLLMs still suffer from structural deficiencies in video review, such as insufficient temporal coverage, leading to extremely high false negative rates in complex scenarios\cite{cao2026failures}. 
To enhance the reliability and logical reasoning ability of video review, multi-agent systems (MAS) have been gradually introduced. MV-Debate framework, for instance, improves the reliability of harmful content detection on social media through a debate mechanism among multiple agents \cite{lu2025mv}, while MAI-SVCR framework integrates multiple specialized agents for minors' protection, national security, etc., for collaborative decision-making \cite{ouyang2025multimodal}. Additionally, memory-augmented agent framework for video understanding, such as VideoAgent \cite{fan2024videoagent}, VideoMultiAgents integrating scene graph analysis \cite{kugo2025videomultiagents}, and the reward-driven self-correction framework ReAgent-V \cite{zhou2025reagent}, have demonstrate potential in handling complex temporal dependencies and logical reasoning in long videos. 
Despite these advancements, most existing research focuses on improving review efficiency or reducing computational costs \cite{wang2025filter,yildirim2024experimentation}, resulting in poor model performance in dealing with malicious or harmful content that highly depends on logical judgment and contextual understanding\cite{ahmed2023malicious,ahmed2023potential}. Transitioning from recognition to deep narrative analysis via multi-agent collaboration remains a critical gap in video logic review.


\begin{figure*}[h]  
  \centering
  \includegraphics[width=1\textwidth]{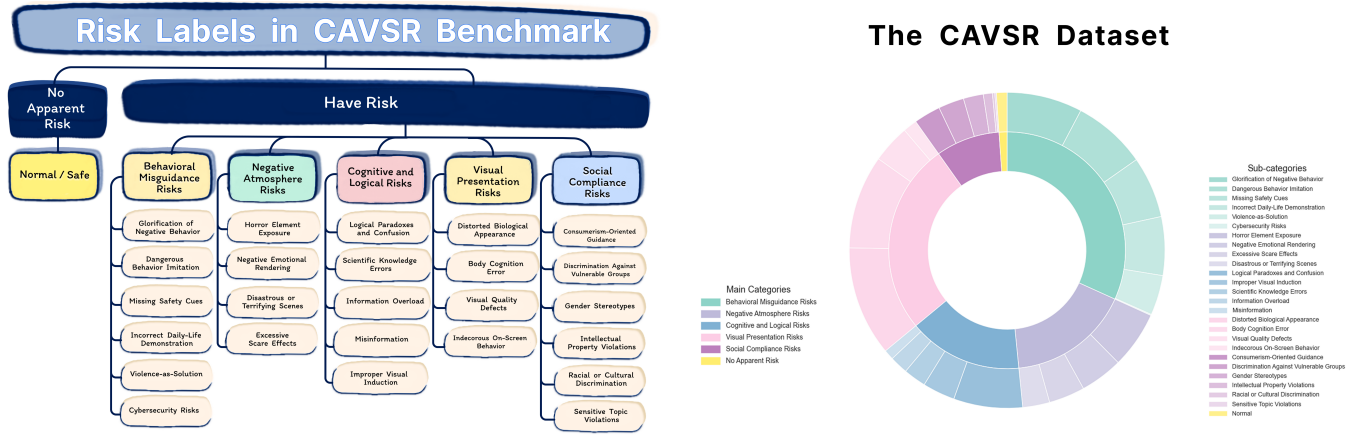} 
  \caption{Overview of the risk label taxonomy in the CAVSR benchmark and the distribution of collected data. The left part illustrates the hierarchical organization of child-oriented AIGC video risks, including 6 top-level categories and 26 fine-grained labels. The right part shows the statistical distribution of the dataset across different risk categories and sub-categories.} 
  \label{fig:database_count}    
\end{figure*} 

\section{The CAVSR Dataset} 
\subsection{Data Preparation}

\paragraph{Collection Criteria}
To construct a dataset for evaluating the suitability of AI-generated videos for children, we collected videos from three mainstream platforms: TikTok, RedNote, and YouTube. We focused on videos explicitly marked as AI-generated. Specifically, candidate videos were identified through two types of cues: (1) the video frames explicitly contained visible markings such as ``AI-generated'' or ``AI''; and (2) the post captions or hashtags included AI-related tags, such as ``AI-generated'' and ``AI video.'' To ensure relevance to children-oriented scenarios, we further constrained the sampling space to videos where AI-related tags co-occurred with child-related tags (e.g., ``children,'' ``early education''). This filtering step allowed us to prioritize content that is more likely to be consumed by or targeted at children.

\paragraph{Data Screening}
The data collection process was conducted manually by four researchers over multiple sessions. Starting from an initial pool of 650 collected videos, we first removed 5 duplicate samples across annotators and platforms. We then excluded 12 videos that contained misleading AI-related tags but were not actually AI-generated. Additionally, 8 videos were removed due to playback issues, corruption, or poor visual quality that hindered subsequent analysis. The final dataset consists of 605 videos for expert annotation and evaluation.

\subsection{Expert Annotation}
\paragraph{Risk Label Classification}


To systematically characterize the risks that children may face when watching AIGC videos, we develop a hierarchical risk taxonomy. Rather than defining labels based on the video topic itself, this taxonomy is organized from the perspective of potential child impact, with particular emphasis on development-sensitive risk factors such as behavioral inducement, emotional stimulation, cognitive misguidance, and value distortion. 
The taxonomy was established by three experts with backgrounds in child education through case analysis and focus group discussion. Specifically, the experts first examined representative AIGC video samples to identify the major types of risk that children may encounter during viewing. Based on these observations, they conducted multiple rounds of focus group discussion to refine the risk boundaries, category organization, and fine-grained definitions, eventually arriving at a complete hierarchical label system. 
As shown in the left part of Figure~\ref{fig:database_count}, the resulting taxonomy consists of six top-level risk categories and 26 fine-grained labels. 
It is important to note that this label system is not used to determine whether a video is for children or not, but rather to identify the potential risks that children may encounter when viewing such content.

\paragraph{Annotation Procedure}

Before formal annotation, the three annotators first received unified guideline training based on the label system, to clarify the definition and boundary of risk labels. The annotators then independently assigned multi-label annotations to all videos. For samples with disagreements, we further organized a review-and-discussion stage, in which the disputed cases were re-examined and discussed until a final consensus was reached.

\subsection{Dataset Analysis}
Overall, the post-annotation label distribution is shown in the right part of Figure~\ref{fig:database_count}. The benchmark exhibits a clear long-tail distribution at both the top-level category and fine-grained label levels. Some risk types, such as \emph{Distorted Biological Appearance}, \emph{Body Cognition Error}, occur relatively frequently in real-world platform videos, whereas others, such as \emph{Cybersecurity Risks}, \emph{Sensitive Topic Violations}, and \emph{Racial or Cultural Discrimination}, are much rarer. 
This pattern is also closely related to the intrinsic characteristics of AIGC videos. High-frequency labels such as \emph{Distorted Biological Appearance} and \emph{Body Cognition Error} can be viewed as direct outcomes of limited generation controllability in biological structure, body proportion, and cross-frame visual consistency. Likewise, the relatively high prevalence of \emph{Dangerous Behavior Imitation}, \emph{Missing Safety Cues}, and \emph{Logical Paradoxes and Confusion} indicates that AIGC videos are prone to producing locally plausible but globally incoherent content, where action logic, safety guidance, and semantic consistency are not always reliably preserved. Negative-atmosphere risks, such as \emph{Horror Element Exposure} and \emph{Negative Emotional Rendering}, may also be amplified by exaggerated synthesis effects and blurred expressive boundaries. Therefore, the dominant labels in our benchmark are not only frequent child-oriented risks on real-world platforms, but also risk types that are especially compatible with the known failure modes of current AIGC video generation.

\section{Methodology}

\begin{figure}[t] 
  \centering
  \includegraphics[width=\linewidth]{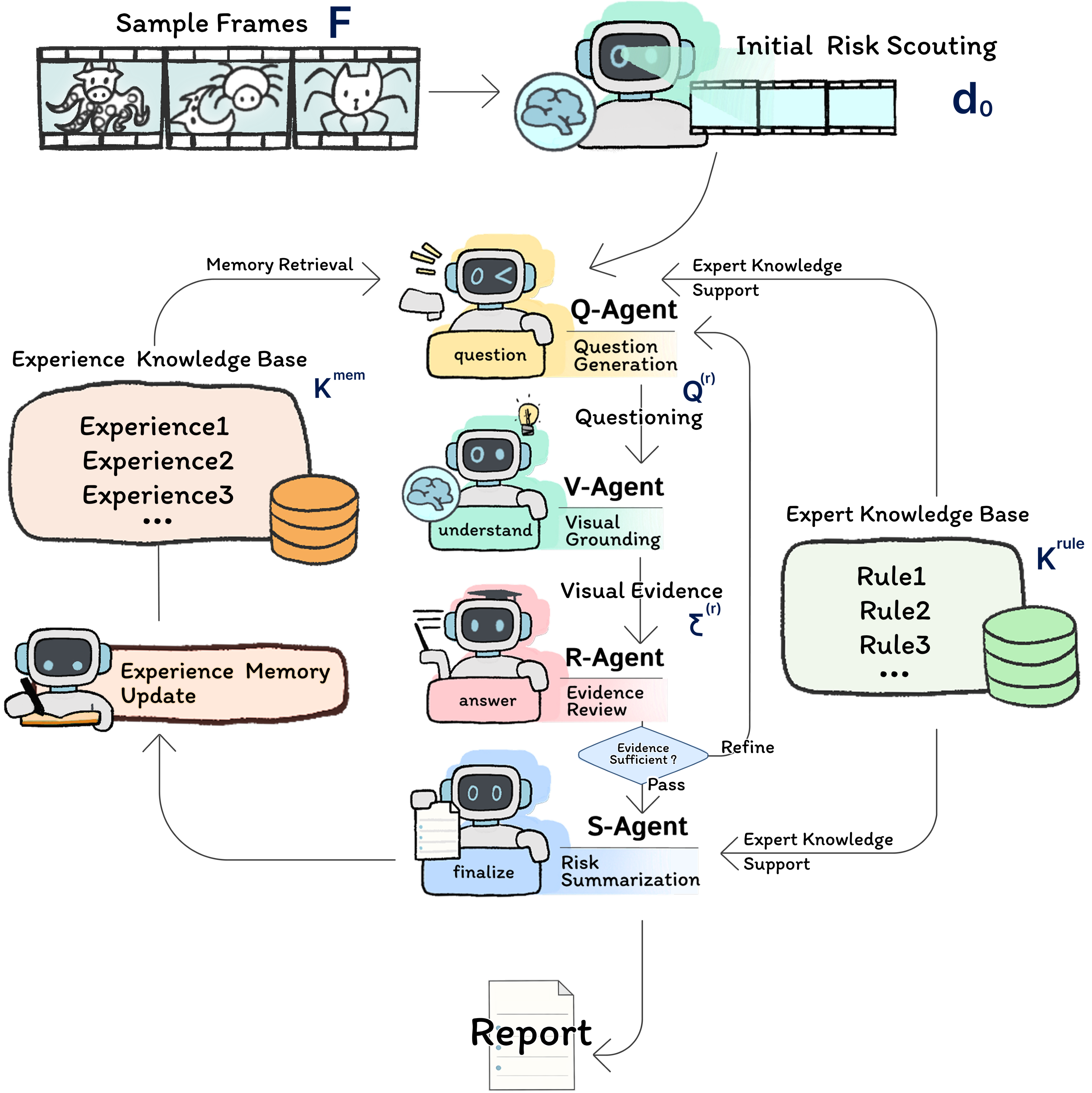}
  \caption{Overview of the proposed knowledge-supported AIGC video review framework (QVRS-E).} 
  \label{fig:method}    
\end{figure}

The AIGC video review faces multiple challenges, including how to reliably identify implicit risks from complex video content, how to avoid missing key evidence in a single round of prediction, and how to effectively utilize expert Knowledge and historical review experience. To address these issues, we propose a knowledge-supported AIGC video review framework, as illustrated in Figure~\ref{fig:method}.
To better capture implicit child-oriented risks, we design a systematic risk evidence construction module, in which risk-guided multi-agent collaboration and iterative Q-V-R probing progressively collect and verify critical visual evidence. We further introduce an experience update and evidence-based assessment module, which writes back reusable reviewing experience into memory and integrates accumulated evidence with expert knowledge to produce more reliable structured risk assessment results.

\subsection{Systematic Risk Evidence Construction}

Our framework consists of four agents: a Question Agent ($\Phi_Q$), a Vision Agent ($\Phi_V$), a Review Agent ($\Phi_R$), and a Summary Agent ($\Phi_S$). $\Phi_Q$, $\Phi_V$, and $\Phi_R$ form an iterative loop for risk evidence collection, while $\Phi_S$ performs the final evidence-based assessment. 

Specifically, $\Phi_Q$ converts potential child-risk cues into targeted verification questions. Given the initial scene description $d_0$ and current feedback, it generates visually verifiable questions targeting risks such as dangerous behaviors, distorted biological appearances, logical anomalies, and frightening atmospheres. $\Phi_V$ serves as the visual verification module, inspecting the sampled frame set $\mathbf{F}$ to return objective answers grounded in visual facts. $\Phi_R$ evaluates whether the currently accumulated evidence $\mathcal{E}^{(r)}$ is sufficient. Given $d_0$ and $\mathcal{E}^{(r)}$, it checks if major risk cues are covered. If evidence is insufficient, it returns a follow-up suggestion; otherwise, it outputs a pass signal. Upon termination of the Q-V-R loop, $\Phi_S$ aggregates the evidence to produce the structured risk report.

\label{sec:evidence_construction}
\subsubsection{Risk-Guided Multi-Agent Design}



\subsubsection{Knowledge-Supported Iterative Q-V-R Probing} 

To overcome the limitations of single-pass prediction and better leverage prior knowledge, we unify expert knowledge, experience knowledge, and iterative Q-V-R probing. Expert knowledge provides rule-level definitions and boundaries, experience knowledge offers reusable historical inquiry patterns, and Q-V-R probing drives progressive evidence construction. Together, they enable targeted questioning and reliable implicit risk assessment.


\paragraph{Expert Knowledge Support}

The expert knowledge base $\mathcal{K}^{\mathrm{rule}}$ consists of expert-authored guidelines covering risk definitions, decision boundaries, typical manifestations, and reviewing considerations. The guidelines are segmented into rule units and stored as vectorized documents in a Chroma database, alongside metadata like risk label tables. Retrieved rule text acts as constraint information for $\Phi_Q$, ensuring generated questions focus on child-related risks while remaining visually verifiable. Further details are provided in the supplementary materials.

\paragraph{Experience Knowledge Enrichment}

The experience knowledge base $\mathcal{K}^{\mathrm{mem}}$ stores historical reviewing memory items as tuples $m=(d, \Pi, y)$, where $d$ is a scene description, $\Pi$ an effective inquiry path, and $y$ the associated risk type. During probing, $\Phi_Q$ uses $d_0$ or the latest feedback to retrieve similar items from $\mathcal{K}^{\mathrm{mem}}$ as references. Since experience knowledge provides priors on questioning strategies rather than direct judgments, $\Phi_Q$ must still generate novel verification questions conditioned on the current video content.

\begin{table*}[t]
  \caption{Performance comparison under the 26 fine-grained labels child-risk recognition setting.}
  \label{tab:results_updated}
  \centering
  \resizebox{\textwidth}{!}{ 
    \setlength{\tabcolsep}{8pt} 
    \begin{tabular}{lcccccccc}
      \toprule
      \textbf{Method} & \textbf{Micro-F1} & \textbf{Macro-F1} & \textbf{Sample-F1} & \textbf{Precision@1} & \textbf{Recall@1} & \textbf{Precision@3} & \textbf{Recall@3} & \textbf{Hit Rate@3} \\
      \midrule
      \multicolumn{9}{l}{\textbf{Open-source Models}} \\
      \midrule
      Qwen2-VL-2B     & 0.0546 & 0.0319 & 0.0991 & 0.1521 & 0.0819 & 0.0650 & 0.0877 & 0.1702 \\
      Qwen2-VL-7B     & 0.1028 & 0.0546 & 0.1130 & 0.3256 & 0.0763 & 0.1262 & 0.0830 & 0.3455 \\
      Qwen2.5-VL-7B   & 0.1246 & 0.0638 & 0.1074 & 0.3091 & 0.0505 & 0.1603 & 0.0718 & 0.3719 \\
      Qwen3-VL-8B     & 0.2414 & 0.1533 & 0.2413 & 0.4843 & 0.1180 & 0.3091 & 0.1811 & 0.5537 \\
      InternVL2.5-8B  & 0.0774 & 0.0434 & 0.1120 & 0.2033 & 0.0848 & 0.0948 & 0.0951 & 0.2083 \\
      GLM-4V-9B       & 0.2805 & 0.1969 & 0.2072 & 0.2711 & 0.0936 & 0.1829 & 0.1320 & 0.3339 \\
      GLM-4.6V-flash  & 0.3030 & 0.2165 & 0.2765 & 0.5174 & \textbf{0.1263} & 0.3366 & 0.1910 & 0.6000 \\
      \midrule
      \multicolumn{9}{l}{\textbf{Our Methods}} \\
      \midrule
      Ours (Qwen2.5-VL-7B) & 0.3208 & 0.1586 & 0.2950 & 0.5504 & 0.1222 & 0.3763 & 0.2075 & 0.6463 \\
      Ours (Qwen3-VL-8B) & \textbf{0.4498} & \textbf{0.2742} & \textbf{0.4108} & \textbf{0.6215} & 0.1221 & \textbf{0.5036} & \textbf{0.2667} & \textbf{0.8264} \\
      \midrule
      \multicolumn{9}{l}{\textcolor{gray}{\textbf{Closed-source Models}}} \\
      \midrule
      \color{gray} GPT-4o      & \color{gray} 0.3898 & \color{gray} 0.2338 & \color{gray} 0.3281 & \color{gray} 0.5615 & \color{gray} 0.1049 & \color{gray} 0.4502 & \color{gray} 0.2195 & \color{gray} 0.6794 \\
      \color{gray} Gemini-3-Flash  & \color{gray} 0.5012 & \color{gray} 0.3462 & \color{gray} 0.4651 & \color{gray} 0.6793 & \color{gray} 0.1395 & \color{gray} 0.5554 & \color{gray} 0.2923 & \color{gray} 0.8430 \\
      \bottomrule
    \end{tabular}
  }
\end{table*}

\paragraph{Iterative Q-V-R Probing}

Given an input video $\mathcal{V}$, we uniformly sample 64 frames $\mathbf{F}=\{f_i\}_{i=1}^{64}$ as the visual basis for reviewing. Rather than directly assigning risk labels, we perform an initial risk scouting step to obtain a coarse scene description $d_0 = \Phi_V(\mathbf{F})$. This description summarizes the main characters, actions, context, and suspicious cues relevant to child-oriented risks.

Based on $d_0$, the framework performs iterative Q-V-R probing. At round $r$, $\Phi_Q$ generates a set of targeted questions:
\begin{equation}
\mathcal{Q}^{(r)}=\Phi_Q\!\left(d_0,\mathcal{E}^{(r-1)},\mathcal{Z}^{(r)}_{\mathrm{mem}},\mathcal{Z}^{(r)}_{\mathrm{rule}}\right),
\end{equation}
where $\mathcal{E}^{(r-1)}$ is the accumulated evidence, $\mathcal{Z}^{(r)}_{\mathrm{mem}}$ the retrieved experience, and $\mathcal{Z}^{(r)}_{\mathrm{rule}}$ the retrieved expert knowledge. $\Phi_V$ then answers these questions over the sampled frames:
\begin{equation}
\mathcal{A}^{(r)}=\Phi_V(\mathbf{F},\mathcal{Q}^{(r)}).
\end{equation}

The resulting question-answer pairs update the evidence set:
\begin{equation}
\mathcal{E}^{(r)}=\mathcal{E}^{(r-1)}\cup\{(q,a)\mid q\in\mathcal{Q}^{(r)},\, a\in\mathcal{A}^{(r)}\}.
\end{equation}
$\Phi_R$ determines whether the evidence is sufficient for assessment:
\begin{equation}
\gamma^{(r)}=\Phi_R(d_0,\mathcal{E}^{(r)}),
\end{equation}
where $\gamma^{(r)}\in\{0,1\}$. If $\gamma^{(r)}=1$, the probing terminates; otherwise, the framework proceeds to the next round.

\subsubsection{Experience Memory Update}
After assessing a video, the system writes newly accumulated experience back to $\mathcal{K}^{\mathrm{mem}}$. Specifically, it extracts the initial description $d_0$, an effective inquiry path $\Pi^\ast$, and the final risk conclusion $y$. Here, $\Pi^\ast$ is represented as a temporally ordered sequence of key question-answer pairs selected from the accumulated evidence $\mathcal{E}^{(R)}$:
\begin{equation}
\Pi^\ast = \big[(q_1,a_1), (q_2,a_2), \ldots, (q_T,a_T)\big],
\end{equation}
where each $(q_t,a_t)$ is a selected question-answer pair from $\mathcal{E}^{(r)}$. The resulting memory item is organized as:
\begin{equation}
m=(d_0,\Pi^\ast,y).
\end{equation}
This memory item is appended to support question generation and inquiry-path reuse in future similar cases:
\begin{equation}
\mathcal{K}^{\mathrm{mem}} \leftarrow \mathcal{K}^{\mathrm{mem}} \cup \{m\}.
\end{equation}
Through this mechanism, the experience knowledge base continuously grows, gradually accumulating more targeted inquiry experience.


\subsubsection{Risk Assessment}
Upon termination of the multi-round probing, the system yields accumulated evidence $\mathcal{E}^{(R)}$, with $R$ being the total executed rounds. Based on this, $\Phi_S$ aggregates the verified Q-A pairs into a unified representation $c=\mathrm{Agg}(\mathcal{E}^{(R)})$, and combines it with relevant expert rules to generate a structured risk report:
\begin{equation}
\mathcal{R}=\Phi_S(c,\mathcal{Z}^{\ast}_{\mathrm{rule}}).
\end{equation}
Instead of a single risk label, the output is a structured report containing a content summary, identified risk points, categories, and potential impacts on children. This ensures the final result preserves core reviewing facts while remaining readable for non-technical users like parents or educators. Because the assessment is grounded in multi-round collection, the report is inherently supported by progressively verified visual facts rather than a one-step classification.

\begin{table*}[t]
  \caption{Performance comparison under the 6-class coarse-grained child-risk recognition setting.}
  \label{tab:results_reordered}
  \centering
  \resizebox{\textwidth}{!}{ 
    \setlength{\tabcolsep}{8pt} 
    \begin{tabular}{lcccccccc}
      \toprule
      \textbf{Method} & \textbf{Micro-F1} & \textbf{Macro-F1} & \textbf{Sample-F1} & \textbf{Precision@1} & \textbf{Recall@1} & \textbf{Precision@3} & \textbf{Recall@3} & \textbf{Hit Rate@3} \\
      \midrule
      \multicolumn{9}{l}{\textbf{Open-source Models}} \\
      \midrule
      Qwen2-VL-2B     & 0.1260 & 0.1315 & 0.1475 & 0.2017 & 0.1117 & 0.0882 & 0.1295 & 0.2198 \\
      Qwen2-VL-7B     & 0.3001 & 0.2249 & 0.2824 & 0.5653 & 0.1902 & 0.2105 & 0.2079 & 0.5752 \\
      Qwen2.5-VL-7B   & 0.3189 & 0.2227 & 0.2770 & 0.5587 & 0.1658 & 0.2275 & 0.1979 & 0.5702 \\
      Qwen3-VL-8B     & 0.4179 & 0.3952 & 0.3862 & 0.5686 & 0.2167 & 0.3135 & 0.3169 & 0.6033 \\
      InternVL2.5-8B  & 0.1147 & 0.1052 & 0.1361 & 0.2132 & 0.1082 & 0.0788 & 0.1137 & 0.2132 \\
      GLM-4V-9B       & 0.3684 & 0.3537 & 0.2903 & 0.3273 & 0.1414 & 0.2275 & 0.2409 & 0.3603 \\
      GLM-4.6V-flash  & 0.4843 & 0.4416 & 0.4288 & 0.6050 & 0.2268 & 0.3631 & 0.3562 & 0.6331 \\
      \midrule
      \multicolumn{9}{l}{\textbf{Our Methods}} \\
      \midrule
      Ours (Qwen2.5-VL-7B) & 0.5176 & 0.4169 & 0.4661 & 0.6512 & 0.2429 & 0.4105 & 0.4075 & 0.6975 \\
      Ours (Qwen3-VL-8B) & \textbf{0.6975} & \textbf{0.5822} & \textbf{0.6474} & \textbf{0.7835} & \textbf{0.2759} & \textbf{0.6154} & \textbf{0.5985} & \textbf{0.8926} \\
      \midrule
      \multicolumn{9}{l}{\textcolor{gray}{\textbf{Closed-source Models}}} \\
      \midrule
      \color{gray} GPT-4o          & \color{gray} 0.5901 & \color{gray} 0.4928 & \color{gray} 0.5014 & \color{gray} 0.6744 & \color{gray} 0.2310 & \color{gray} 0.4596 & \color{gray} 0.4226 & \color{gray} 0.7126 \\
      \color{gray} Gemini-3-Flash  & \color{gray} 0.7516 & \color{gray} 0.6768 & \color{gray} 0.7008 & \color{gray} 0.8132 & \color{gray} 0.2928 & \color{gray} 0.6501 & \color{gray} 0.6350 & \color{gray} 0.9124 \\
      \bottomrule
    \end{tabular}
  }
\end{table*}

\section{Experiments}

\subsection{Experimental Settings}
We compare the proposed method with a set of representative open- and closed-source multimodal large models, including Qwen2-VL \cite{wang2024qwen2}, Qwen2.5-VL \cite{wang2024qwen2}, Qwen3-VL \cite{bai2025qwen3}, InternVL2.5 \cite{chen2024expanding}, GLM-4V \cite{glm2024chatglm}, GPT-4o \cite{hurst2024gpt}, and Gemini-3-Flash \cite{google2025gemini3flash}. 
To evaluate both fine-grained and coarse-grained child-risk recognition, we follow the multi-label evaluation setting adopted in prior work~\cite{bogatinovski2022comprehensive,he2024representation} and report results under both the 26 fine-grained setting and the merged 6-category setting, using Micro-F1, Macro-F1, Sample-F1, as well as Precision@K and Recall@K. In addition, to better reflect whether the model can successfully retrieve at least one correct risk label within the top-$K$ predictions, we further introduce Hit@K as an auxiliary ranking metric. The formal definition of Hit@K is provided in the supplementary material.

Among the evaluated models, all open-source backbones are within the sub-10B parameter scale, including Qwen2-VL-2B, Qwen2-VL-7B, Qwen2.5-VL-7B, Qwen3-VL-8B, InternVL2.5-8B, GLM-4.6V-flash and GLM-4V-9B. 
In our implementation, all four agents use the same VLM backbone, either Qwen2.5-VL-7B or Qwen3-VL-8B. Both the experience knowledge base and the expert knowledge base are implemented with Chroma, using m3e-base as the text embedding model. All experiments are conducted on a single NVIDIA A100 GPU with 80GB memory.


\subsection{Evaluation Results}
To evaluate the effectiveness of the proposed framework, we conduct comprehensive experiments on both the 26-class fine-grained setting and the 6-class setting. The results are reported in Tables~\ref{tab:results_updated} and~\ref{tab:results_reordered}. Based on these results, we have following observations:

\begin{itemize}[leftmargin=*, nosep]
\item \textbf{The proposed framework consistently improves different open-source VLM backbones, showing that the gains come from the framework design rather than a single backbone choice.} 
As shown in Tables~\ref{tab:results_updated} and~\ref{tab:results_reordered}, both Qwen2.5-VL-7B and Qwen3-VL-8B benefit substantially from the proposed framework under the fine-grained and coarse-grained settings. Under the 26-class setting, Micro-F1 increases from 0.1246 to 0.3208 on Qwen2.5-VL-7B and from 0.2414 to 0.4498 on Qwen3-VL-8B, with similarly clear gains on top-$K$ ranking metrics. Under the 6-class setting, the same trend remains, where the Qwen3-VL-8B-based variant further reaches 0.6975 on Micro-F1 and 0.8926 on Hit@3. Since both backbones show clear improvements in both settings, these gains are more plausibly attributed to the proposed multi-round Q-V-R probing and knowledge-supported reviewing process, rather than to a particular backbone alone. 

\item \textbf{GLM and Qwen series models exhibit different strengths in fine-grained risk recognition, while our framework leads to more balanced gains across classification and ranking metrics.} 
The results show that the GLM and Qwen series emphasize different aspects of performance. Under the 26-class setting, GLM-4.6V-flash achieves the highest Recall@1 among open-source models, reaching 0.1263, suggesting that it is relatively strong at ranking one true label early. However, Ours (Qwen3-VL-8B) performs substantially better on the other major metrics, including Micro-F1, Macro-F1, Precision@3, Recall@3, and Hit@3. A similar pattern can also be observed under the 6-class setting. GLM-4.6V-flash remains a relatively strong open-source baseline, with Micro-F1 of 0.4843 and Recall@3 of 0.3562, but Ours (Qwen3-VL-8B) further raises these values to 0.6975 and 0.5985, respectively, while also achieving the best Precision@1 and Hit@3 among all open-source models. These results suggest that some model families may be stronger on isolated ranking indicators, but our framework produces more balanced gains across both classification quality and top-$K$ ranking quality, which is more aligned with the requirement of child-oriented reviewing, where the model must not only detect whether risk exists, but also organize multiple risk labels in a more reliable order. 

\item \textbf{With the proposed framework, a 7B--8B open-source backbone already becomes highly competitive with strong closed-source models, showing practical value for local deployment and low-cost iteration.} 
Although closed-source models remain strong overall, the proposed framework substantially narrows the gap between medium-scale open-source backbones and proprietary systems. Under the 26-class setting, Ours (Qwen3-VL-8B) achieves Micro-F1 of 0.4498, Macro-F1 of 0.2742, and Hit@3 of 0.8264, all higher than the corresponding GPT-4o results. Under the 6-class setting, the same model reaches 0.6975 on Micro-F1 and 0.8926 on Hit@3, again exceeding GPT-4o, whose corresponding results are 0.5901 and 0.7126. Although Gemini-3-Flash remains the strongest model overall, the remaining gap is already small on some coarse-grained metrics: for example, Ours (Qwen3-VL-8B) reaches 0.7835 on Precision@1 and 0.8926 on Hit@3, while Gemini-3-Flash reaches 0.8132 and 0.9124. Considering that closed-source models generally enjoy stronger advantages in model scale, system optimization, and inference resources, these results indicate that applying the proposed framework to 7B--8B open-source backbones already yields strong practical performance and shows promising potential for local deployment and low-cost iteration. 

\item \textbf{The proposed framework becomes even more effective when the label space is semantically more stable, indicating that the accumulated evidence can be translated more reliably into high-level child-risk judgments.}
Compared with the 26 fine-grained setting, the gains brought by our framework are generally larger and more stable under the merged 6-class setting. For Qwen3-VL-8B, Micro-F1 increases by 0.2084 under the 26-class setting, but by 0.2796 under the 6-class setting; Recall@3 increases by 0.0856 in the fine-grained setting, but by 0.2816 in the coarse-grained setting. For Qwen2.5-VL-7B, Micro-F1 increases by 0.1962 under the 26-class setting and by 0.1987 under the 6-class setting, while Recall@3 increases from 0.0718 to 0.2075 in the former and from 0.1979 to 0.4075 in the latter. This pattern suggests that when the target label space is merged into more semantically stable high-level categories, the proposed framework can more effectively convert iterative evidence construction and knowledge support into reliable child-risk recognition results.

\end{itemize}

\subsection{Ablation Study}

To analyze the contributions of different components in the proposed framework, we conduct ablation experiments by removing iterative Q-V-R probing and Expert Knowledge Support. The results under both the 26-class and 6-class settings are shown in Tables~\ref{tab:ablation_26} and~\ref{tab:ablation_6}. We have following findings:
 
\begin{itemize}[leftmargin=*, nosep]
\item \textbf{Iterative Q-V-R probing is the primary source of performance gain.} As shown in Tables~\ref{tab:ablation_26} and~\ref{tab:ablation_6}, removing iterative Q-V-R probing leads to the largest overall degradation under both settings. Under the 26-class setting, the full model improves Micro-F1 from 0.4321 to 0.4498 and Hit@3 from 0.7603 to 0.8264 over the probing-removed variant. Under the 6-class setting, Precision@1 and Hit@3 further increase from 0.7603 to 0.7835 and from 0.8595 to 0.8926. These results suggest that multi-round probing mainly improves evidence coverage by turning one-step judgment into a progressive evidence construction process. 

\begin{table}[t]
  \caption{Ablation study of the proposed framework under the 26 fine-grained setting. }
  \label{tab:ablation_26}
  \centering
  \resizebox{\columnwidth}{!}{ 
    \setlength{\tabcolsep}{2pt}
    \begin{tabular}{lcccccccc}
      \toprule
      Variants / Dimension & Mi-F1 & Ma-F1 & S-F1 & P@1 & R@1 & P@3 & R@3 & Hit@3 \\
      \midrule
      \textbf{Baseline}: Qwen3-VL-8B & 0.2414 & 0.1533 & 0.2413 & 0.4843 & 0.1180 & 0.3091 & 0.1811 & 0.5537 \\
      \addlinespace[0.3em]
      \textbf{Baseline} + Expert Knowledge            & 0.3559 & 0.1899 & 0.3237 & 0.4579 & 0.1079 & 0.4248 & 0.2238 & 0.6992 \\
      \textbf{Baseline} + Iterative Q-V-R Probing           & 0.4321 & 0.2495 & 0.3800 & 0.5372 & 0.0976 & 0.4325 & 0.2157 & 0.7603 \\
      \midrule
      \textbf{Ours} (all)                & \textbf{0.4498} & \textbf{0.2742} & \textbf{0.4108} & \textbf{0.6215} & \textbf{0.1221} & \textbf{0.5036} & \textbf{0.2667} & \textbf{0.8264} \\
      \bottomrule
    \end{tabular}
  }
\end{table}

\begin{table}[t]
  \caption{Ablation study of the proposed framework under the 6-class coarse-grained setting. }
  \label{tab:ablation_6}
  \centering
  \resizebox{\columnwidth}{!}{ 
    \setlength{\tabcolsep}{2pt} 
    \begin{tabular}{lcccccccc}
      \toprule
      Variants / Dimension & Mi-F1 & Ma-F1 & S-F1 & P@1 & R@1 & P@3 & R@3 & Hit@3 \\
      \midrule
      \textbf{Baseline}: Qwen3-VL-8B & 0.4179 & 0.3952 & 0.3862 & 0.5686 & 0.2167 & 0.3135 & 0.3169 & 0.6033 \\
      \textbf{Baseline} + Expert Knowledge            & 0.6791 & \textbf{0.6242} & 0.5964 & 0.6727 & 0.2442 & 0.5620 & 0.5274 & 0.7752 \\
      \textbf{Baseline} + Iterative Q-V-R Probing           & 0.6960 & 0.5726 & 0.6322 & 0.7603 & 0.2621 & 0.6055 & 0.5725 & 0.8595 \\
      \midrule
      \textbf{Ours} (all)                & \textbf{0.6975} & 0.5822 & \textbf{0.6474} & \textbf{0.7835} & \textbf{0.2759} & \textbf{0.6154} & \textbf{0.5985} & \textbf{0.8926} \\
      \bottomrule
    \end{tabular}
  }
\end{table}

\begin{figure*}[t] 
  \centering
  \includegraphics[width=0.98\textwidth]{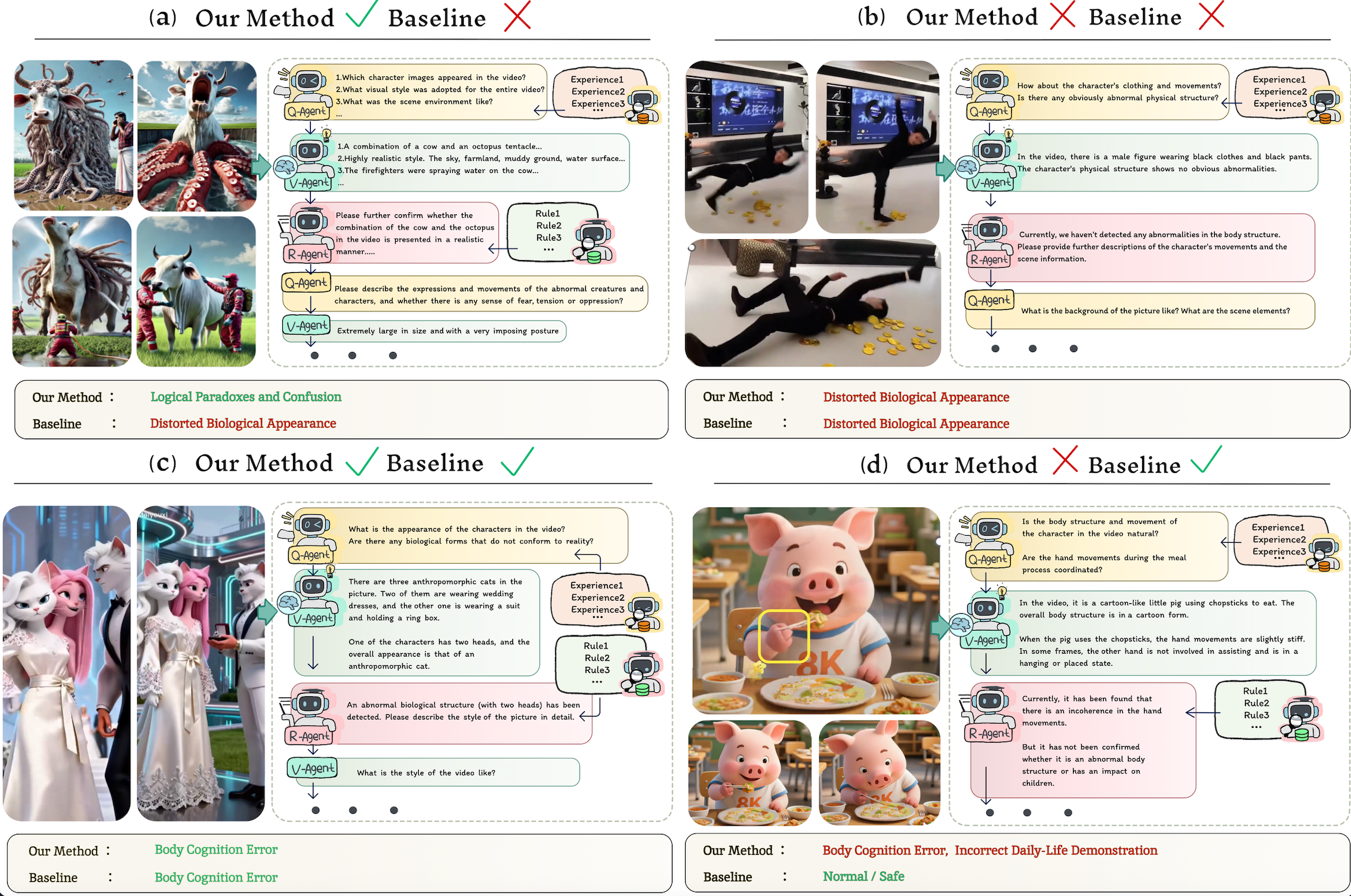} 
  \caption{Representative cases in the error analysis. Cases (a)--(d) compare our method with the baseline on child-oriented risk recognition in AIGC video review, where $\checkmark$ and $\times$ denote correct and incorrect predictions, respectively. Green labels indicate correct predictions, and red labels indicate incorrect ones. Only partial Q--V--R interactions are shown for illustration.} 
  \label{fig:error}    
\end{figure*}

\item \textbf{Expert knowledge support provides additional gains, especially for fine-grained risk and label ranking.} The variant with Expert knowledge alone already outperforms the backbone under both settings, but still remains weaker than the variant with iterative Q-V-R probing alone on most major metrics. Under the 26-class setting, adding expert knowledge raises Micro-F1 from 0.2414 to 0.3559 and Recall@3 from 0.1811 to 0.2238. This indicates that expert knowledge does not replace evidence construction, but mainly helps the model distinguish related risk categories more reliably and improve fine-grained label ranking. 

\item \textbf{The two components are complementary, although their contributions are reflected differently across metrics and label granularities.} In the 26-class setting, adding expert knowledge on top of iterative probing brings clear gains, with Macro-F1 increasing from 0.2495 to 0.2742 and Recall@3 increasing from 0.2157 to 0.2667. In the 6-class setting, the additional gain becomes smaller, and the probing-only variant attains a slightly higher Macro-F1 than the full model. We note that Macro-F1 is a class-balanced metric, and its slight fluctuation does not contradict the stronger overall performance of the full model on the other major metrics. Taken together, these results suggest that Q-V-R probing mainly helps the model find more complete evidence, while expert knowledge mainly helps the model interpret the evidence more reliably, especially in fine-grained settings.

\end{itemize}

\subsection{Error Analysis}

Figure~\ref{fig:error} compares our method with the baseline in typical cases to demonstrate its applicability and analyze specific failure modes.

\textbf{Explicit Risks:} When visual risk evidence is explicit and stable across frames, both methods generally succeed. In case~(c), the two-headed anthropomorphic cat remains clearly visible in the main subject region, making the abnormal biological structure easily detectable for both models. \textbf{Implicit/Logical Risks:} Crucially, our method excels when risks are implicit and depend on logical or cognitive interpretation. Case~(a) depicts an unrealistic hybrid creature embedded in a realistic human-participating scene. The key risk lies in the realistic framing of an impossible event, which is difficult to identify through single-pass global judgment. While the baseline focuses on surface-level visual cues, our framework progressively narrows down suspicious elements via Q-V-R interaction to successfully uncover deeper logical risks. \textbf{Transient/Fine-grained Risks:} Both methods can fail when abnormal cues appear only briefly or depend on local fine-grained details. In case~(b), the critical cue flashes momentarily and is not stably preserved in the sampled frames. Such cases are difficult because evidence is missed at the perception stage, primarily due to uniform frame sampling, limited viewpoints, or insufficient spatial resolution. \textbf{Over-interpretation:} Our failure typically occurs when benign or stylized content contains weak anomalies that are over-interpreted. For instance, in case~(d), our model misinterprets a cartoon pig's stiff movements as abnormal behavior due to the Q-Agent's strong risk-oriented prior, demonstrating that our framework can be overly conservative in stylized content compared to the baseline.

\section{Conclusion}
In this work, we construct a child-oriented benchmark based on real-world platform videos, together with a hierarchical risk taxonomy with 6 top-level risk categories and 26 fine-grained labels. Building on the benchmark, we further propose a knowledge-supported AIGC video review framework, in which multi-round Q-V-R probing, combining with expert and experience knowledge, is used to support progressive evidence construction and structured risk assessment. Experimental results show that the proposed method consistently improves child-oriented risk recognition under both the fine-grained 26-class setting and the coarse-grained 6-class setting. 
We hope the proposed benchmark will support future child-oriented video reviewing research, while QVRS-E can serve as a robust and extensible baseline for open-source models. Future work will examine its effectiveness across different model scales and improve multi-agent efficiency for real-time deployment.

\bibliographystyle{ACM-Reference-Format}
\bibliography{sample-base}

\appendix

\end{document}